\documentclass{article}

\usepackage{arxiv}
\usepackage{algorithm}
\usepackage{algorithmic}
\usepackage{float}
\usepackage[utf8]{inputenc} 
\usepackage[T1]{fontenc}    
\usepackage{hyperref}       
\usepackage{url}            
\usepackage{booktabs}       
\usepackage{amsfonts}       
\usepackage{nicefrac}       
\usepackage{microtype}      
\usepackage{amsmath}
\usepackage{lipsum}
\usepackage{graphicx}
\usepackage{nth}
\begin{document}
\title{The Importance of Robust Features in Mitigating Catastrophic Forgetting}

\author{
Hikmat Khan \\
\textit{Dept. of Electrical and }\\ \textit{Computer Engineering} \\
\textit{Rowan University}\\
Glassboro, New Jersey, USA \\
bouaynaya@rowan.edu
\And
Nidhal C. Bouaynaya\\
\textit{Dept. of Electrical and }\\ 
\textit{Computer Engineering} \\
\textit{Rowan University}\\
Glassboro, New Jersey, USA \\
bouaynaya@rowan.edu
\And
Ghulam Rasool\\
\textit{Dept. of Machine Learning} \\
\textit{Moffitt Cancer Center}\\
Tampa, Florida, USA \\
ghulam.rasool@moffitt.org\\
}
\maketitle    
\begin{abstract}

Continual learning (CL) is an approach to address catastrophic forgetting, which refers to forgetting previously learned knowledge by neural networks when trained on new tasks or data distributions. The adversarial robustness has decomposed features into robust and non-robust types and demonstrated that models trained on robust features significantly enhance adversarial robustness. However, no study has been conducted on the efficacy of robust features from the lens of the CL model in mitigating catastrophic forgetting in CL. In this paper, we introduce the CL robust dataset and train four baseline models on both the standard and CL robust datasets. Our results demonstrate that the CL models trained on the CL robust dataset experienced less catastrophic forgetting of the previously learned tasks than when trained on the standard dataset. Our observations highlight the significance of the features provided to the underlying CL models, showing that CL robust features can alleviate catastrophic forgetting.


\end{abstract}

\section{Introduction}
Continual learning (CL), also known as lifelong or incremental learning, is a critical component of artificial intelligence (AI). Its primary role is to facilitate the acquisition of new knowledge by learning systems while preserving previously acquired knowledge. CL model's ability to continually learn and build upon past knowledge is essential in scenarios where data distribution is subject to change over time, enabling AI systems to remain adaptable and flexible. Deep learning methods have shown remarkable performance in various tasks but are plagued by a significant challenge known as catastrophic forgetting \cite{de2021continual,  mccloskey1989catastrophic}. This phenomenon occurs when the models become too fixated on new data, leading to forgetting previously acquired knowledge. Such a challenge poses a fundamental tradeoff between plasticity and stability in the CL models \cite{mermillod2013stability}. The CL model must be plastic enough to learn from new data and stable enough to retain previous knowledge \cite{mermillod2013stability}. Balancing plasticity and stability is challenging as data distribution and feature importance change with new tasks, leading to poor performance on previous ones \cite{de2021continual}. Solving the plasticity-stability dilemma is crucial for developing deep learning methods that can continually learn and adapt without forgetting previous knowledge.

Various approaches have been proposed to tackle the challenge of catastrophic forgetting in CL. \textbf{Regularization-based approaches} aim to maintain a balance between old and new tasks by controlling the changes to the critical parameters of the network \cite{aljundi2018memory, kirkpatrick2017overcoming, lee2017overcoming, zenke2017continual}. \textbf{Structure-based approaches} utilize the network structure to partition parameters and prevent interference between tasks \cite{mallya2018packnet, fernando2017pathnet, yoon2017lifelong, li2019learn, rusu2016progressive}. They add new modules to the network while keeping previously learned modules fixed, allowing the model to incorporate new knowledge while retaining old information \cite{fernando2017pathnet,  douillard2022dytox, golkar2019continual, hung2019compacting, collier2020routing, wen2020batchensemble}. \textbf{Replay-based approaches} store and replay data from previous tasks during training \cite{lopez2017gradient, chaudhry2019tiny, rebuffi2017icarl, rolnick2019experience, aljundi2019gradient, isele2018selective, liu2020generative, shin2017continual, atkinson2018pseudo, lavda2018continual}. While learning a new task, previous examples are rehearsed in their original form or pseudo-generated samples using a separately trained generator \cite{rebuffi2017icarl, rolnick2019experience, aljundi2019gradient, isele2018selective, liu2020generative, shin2017continual, atkinson2018pseudo, lavda2018continual}. Replay-based approaches have yielded state-of-the-art results in CL. \textbf{Knowledge distillation-based} approaches aim to reduce the dissimilarity between the representations of a preceding model and the current one while also allowing for efficient adaptation to new tasks \cite{kang2022class, jung2016less, li2017learning}. Recently, \textbf{Dataset distillation-based approaches}, which involve compressing the knowledge learned from previous tasks into a smaller subset of the original dataset, gained considerable attention \cite{rosasco2022distilled, deng2022remember}. The examples from the smaller subset of the old tasks are then replayed when the model is trained on the new task. The distilled knowledge obtained from prior tasks has been shown to exhibit memory efficiency and effectively retain the knowledge learned from previous tasks \cite{rosasco2022distilled}.

The concept of robust and non-robust features in adversarial learning has been proposed in the literature \cite{ilyas2019adversarial, engstrom2019adversarial}. Robust features resist adversarial attacks and help a model maintain accuracy in the presence of adversarial attacks. In contrast, non-robust features are vulnerable to such attacks, resulting in erroneous predictions. This distinction has been leveraged to construct robust datasets, and studies have demonstrated that using robust features significantly improves the adversarial robustness of a model \cite{engstrom2019adversarial, ilyas2019adversarial,waqas2021brain, khan2020explainable, waqas2022exploring, waqas2023multimodal, khan2021deep} against adversarial attacks. The utilization of robust features is an active area of research in adversarial learning, as it can enhance the security and reliability of the model in various applications, including autonomous driving, healthcare, and finance. To our knowledge, no studies have been conducted to evaluate whether providing the CL model with a CL robust dataset or features mitigates catastrophic forgetting. This gap in the literature represents an important avenue for further research, as understanding the potential benefits of CL robust datasets or features contributes to the development of more adaptable CL models capable of retaining previously learned knowledge while adapting to new knowledge.

This paper presents a novel inquiry into the critical role of CL robust features in training CL models, explicitly focusing on class incremental learning (CIL) using the CIFAR10 dataset. The study aims to explore the impact of the CL robust features on the performance of CL models in mitigating catastrophic forgetting. To create the robustified dataset, we used the replay strategy to train the CL model, which obtained the highest average accuracy (referred to as the ``oracle model"). We subsequently used the oracle model to create the robustified version of the CIFAR10 dataset, referred to as the CL robust dataset ($\mathcal{D_R}$). Our experimental results show that training the CL models on the $\mathcal{D_R}$: i) leads to higher average accuracy, and ii) results in less catastrophic forgetting of the previous task. Our study's results agree with previous research in the field of adversarial learning, which has demonstrated the effectiveness of training models on robust features in enhancing their resilience to adversarial attacks. Specifically, we observed that training the CL models on $\mathcal{D_R}$ reduced the occurrence of catastrophic forgetting of the previously learned task. Further, these findings suggest that using $\mathcal{D_R}$ mitigates catastrophic forgetting in CL, highlighting the potential benefits of incorporating such robust features into the training of CL models.  

This paper is organized as follows. Section \ref{sec:methodology} provides a comprehensive account of the methodology employed in training the oracle model and creating the $\mathcal{D_R}$. Section \ref{sec:experimental_setup} describes the experimental setup. Section \ref{sec:result_and_discussion} presents a detailed analysis and discussion of the experimental results. Lastly, Section \ref{sec:conclusion} summarizes the key observations and concludes the paper.

\section{Continual Learning Robust Features}
\label{sec:methodology}
\subsection{Hypothesis and Experimental Design}
One explanation for catastrophic forgetting is that the CL model's features, or the representations of the input data that the model uses, are not sufficiently generalizable to new tasks\cite{de2021continual}. When the model is trained on a new task, it updates its feature representations to fit the new data better but causes the model to forget its previous knowledge \cite{yu2020semantic}. We hypothesize that the features extracted by a deep learning model cause catastrophic forgetting. To test this hypothesis, we propose an experimental study where we disentangle CL robust features from non-robust features in benchmark image classification datasets. CL non-robust features (or spurious features) may be important for accuracy but are not generalizable across tasks \cite{lesort2022continual}. Specifically, given a training dataset, we propose an experimental design to extract a ``CL robustified" ($\mathcal{D_R}$) version of the dataset for CL robust classification, mitigating catastrophic forgetting. 

\subsection{Creating CL Robust Dataset}
\label{sec:Create_CL_Dataset}
Let $f_t$ be a model and $f_t^o$ be an oracle CL robust model at time $t$. We define a feature $f$ as a function mapping from the input space to the real numbers, with the set of all features denoted by $\mathcal{F}$. Given a standard training dataset ($\mathcal{D}$), we would like to construct a $\mathcal{D_R}$ such that

\begin{equation} \label{eq1}
\resizebox{0.5\textwidth}{!}{
    $E_{(x,y) \sim D_R} [f_t(x_{t-\tau}).y_{t-\tau}] = \begin{cases}
    E_{(x,y) \sim D} [f^o_t(x_{t-\tau}).y_{t-\tau}], \text{if } f \in F^o, \tau = 1,\cdots,t. \\
    0, \text{otherwise}
    \end{cases}$
}
\end{equation}

Equation (\ref{eq1}) states that the $\mathcal{D_R}$ is created such that the model at time $t$ does not forget the training data (i.e., task information) at previous times $t-\tau, \tau = 1,\cdots,t$. 

The oracle model is used to carry out the extraction of CL robust features for $\mathcal{D_R}$. This process involves extracting the task-specific knowledge acquired by the oracle model. The resulting $\mathcal{D_R}$ will comprise the highlighted or extracted features that ease the learning of the CL models. The extraction process enhances the efficiency and effectiveness of the CL models by providing them with more focused and relevant CL robust features (i.e., CL robust version). Formally, the optimization objective for  CL feature robust extraction utilizing the oracle model can be expressed as follows:

\begin{equation} 
    x_{cl} = \operatorname*{argmin}_{x_{z}\in[0,1]^d}\| 
    f^{o}_{{t}}(x_{z}) 
   - f^{o}_{{t}}(x_{t-\tau})\|^{2}_{2}, \tau = 1, \cdots,t.
\label{eq:x_cl}
\end{equation}
where $f^{0}_{t}$ is the oracle model at time step $t$, $x_{t-\tau}$ is the target sample, and $x_{cl}$ is the CL robust sample. Equation (\ref{eq:x_cl}) is solved numerically using stochastic gradient descent.

\begin{figure*}[htbp]
  \centering
  \includegraphics[width=0.95\linewidth]{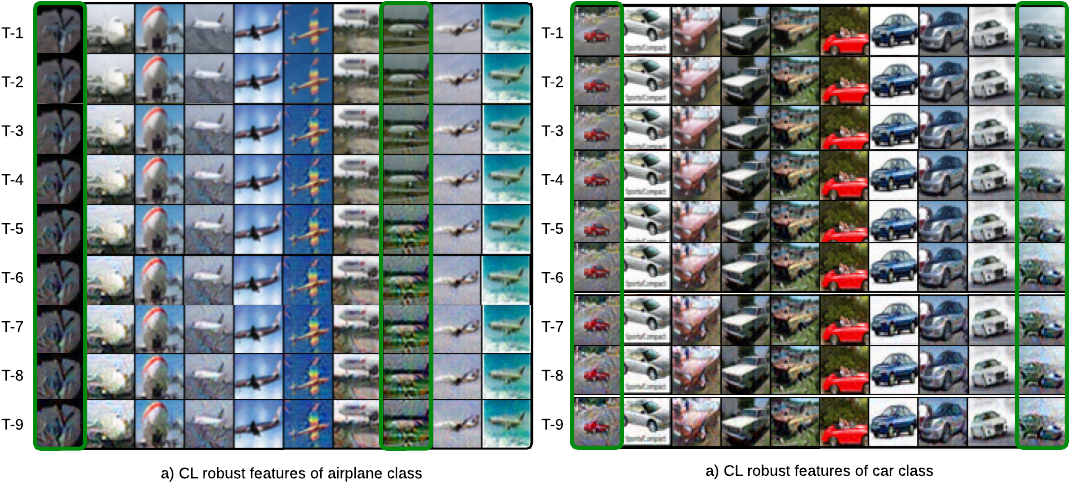}
  \caption{Sub-figures a) and b) illustrate the task-wise CL robust features for the airplane and car class, respectively. The top row  (i.e., T-1) shows original images, while the subsequent rows display CL robust features learned by the oracle model after each task. The CL robust features exhibit continuous evolution as the model learns each task sequentially, as evident from the changing features (in the image) in subsequent rows. The task-specific CL robust features for the airplane and car class, as identified by the oracle CL model, can be readily observed by examining the first and eight columns in sub -figure a) and the first and last column in sub-figure b)  enclosed within a green rectangular (Best viewed in color).
  }
  \label{fig:fig_cl_features}
\end{figure*}

To understand Eq. (\ref{eq:x_cl}), consider the example of learning nine tasks sequentially. The first task consists of two classes; subsequent tasks comprise one class per task. The oracle model learns the first task at the time $t=1$. Then, the oracle model learns the second task at the time $t=2$. Before learning the third task, the CL robust features for the learned classes at the time  $t=1$ are extracted using Eq. (\ref{eq:x_cl}). Upon completion of each task, CL robust features extraction is performed for the learned tasks to compensate for any drift in features and update the relative importance of features that occur after learning new tasks. Figure \ref{fig:fig_cl_features} presents the CL robust features specific to airplane class. The CL robust samples (i.e., $x_{cl}$) are constructed via the mapping $x_{t-\tau} \rightarrow x_{cl}$ using $f^{o}_{t}$ and added to the CL robust dataset along with their corresponding label $y_{t-\tau}$. We provide the pseudo-code for the creation of the $\mathcal{D}_{R}$ in
Algorithm \ref{alg:f_distill}. The algorithm \ref{alg:f_distill} takes the oracle model $f^{o}$, $\mathcal{D}$, $\mathcal{M}$, and learning rate $\alpha$ as input parameters. It outputs the dataset $\mathcal{D_{R}}$. In step 1, $\mathcal{D_{R}}$ is initialized as an empty set. In step 2, the algorithm iterates over the $\mathcal{D}$ and selects each data point $x$ and its corresponding ground truth label $y$. Step 3 performs a minimization operation for $z$ to obtain the final $x_{cl}$ value. Finally, in step 4, the created CL robust sample, represented by $x_{cl}$ and the corresponding ground truth $y$, is added to the  $\mathcal{D_{R}}$. 

\subsection{Training the Oracle CL Model}
\label{sec:oracle_model}
We employed a replay-based continual learning strategy to obtain the oracle model. However, it is worth noting that other CL strategies could also be used to obtain the oracle model. To create the $\mathcal{D}_{R}$, we trained the oracle CL model using the replay-based strategy \cite{de2021continual} with a buffer size of 1000 to store previous task examples. Algorithm \ref{algo:replay} outlines the steps involved in training the oracle model, where $\mathcal{M}$ refers to the memory buffer, $\mathcal{T}$ denotes the total number of sequential tasks, $f_{\theta}$ represents the CL model, and $\alpha$ denotes the learning rate utilized during the training process. Specifically, the algorithm $\ref{algo:replay}$ trains model $f_{\theta}$ using dataset $\mathcal{D}$, tasks $\mathcal{T}$, memory $\mathcal{M}$, and learning rate $\alpha$, resulting in the trained model $f_{\theta}^{*}$. It iterates over tasks, samples the current task batch, then concatenates previous task data/memories to the current batch, performs learning, and in the last step, it adjusts the memory by adding the subset of the current task data.

\begin{algorithm}
 \caption{Replay Based Training}
 \begin{algorithmic}[1]
 \renewcommand{\algorithmicrequire}{\textbf{Input:}}
 \renewcommand{\algorithmicensure}{\textbf{Output:}}
 
 \REQUIRE {$\mathcal{D}, \mathcal{T}, \mathcal{M}, f_{\theta}, \alpha$}
 
 \ENSURE  {$f^{*}_{\theta}$}

  \FOR{$t\in\{1,\dots,\mathcal{T}\}$}
  
        \FOR{$B_{n}\sim\mathcal{D}_t$}
        
        \STATE $B_{\mathcal{M}_{t}} \leftarrow MemoryRetrival(B_{n}, \mathcal{M})$
        
        \STATE $\theta \leftarrow SGD(B_{n}\cup B_{\mathcal{M}_{t}}, \theta, \alpha)$
        \STATE $B_{\mathcal{M}_{t}} \leftarrow MemoryUpdate(B_{n}, \mathcal{M})$
        
        \ENDFOR

  \ENDFOR
 \end{algorithmic} 
 \label{algo:replay}
 
\end{algorithm}

\begin{algorithm}

 \caption{CL Robust Dataset ($\mathcal{D_R}$)}
 \begin{algorithmic}[1]
 \renewcommand{\algorithmicrequire}{\textbf{Input:}}
 \renewcommand{\algorithmicensure}{\textbf{Output:}}
 
 \REQUIRE {$ f^{o},\mathcal{D},  \mathcal{M}, \alpha$}
 
 \ENSURE  {$\mathcal{D}_{R}$}
 

  \STATE $\mathcal{D}_{R} \leftarrow \{\}$ 

  \FOR{$(x, y)\sim\mathcal{D}$}

    \STATE $x_{cl} \leftarrow \operatorname*{argmin}_{z\in[0,1]^d}\| f^{o}(z) - f^{o}(x)\|^{2}_{2}$ 
    
    \STATE $\mathcal{D}_{R} \leftarrow \mathcal{D}_{R} \cup\{(x_{cl}, y)\}$
      
    \ENDFOR
  
 \end{algorithmic} 
\label{alg:f_distill}
\end{algorithm}

\begin{figure*}[htbp]
  \centering
  \includegraphics[width=0.90\linewidth]{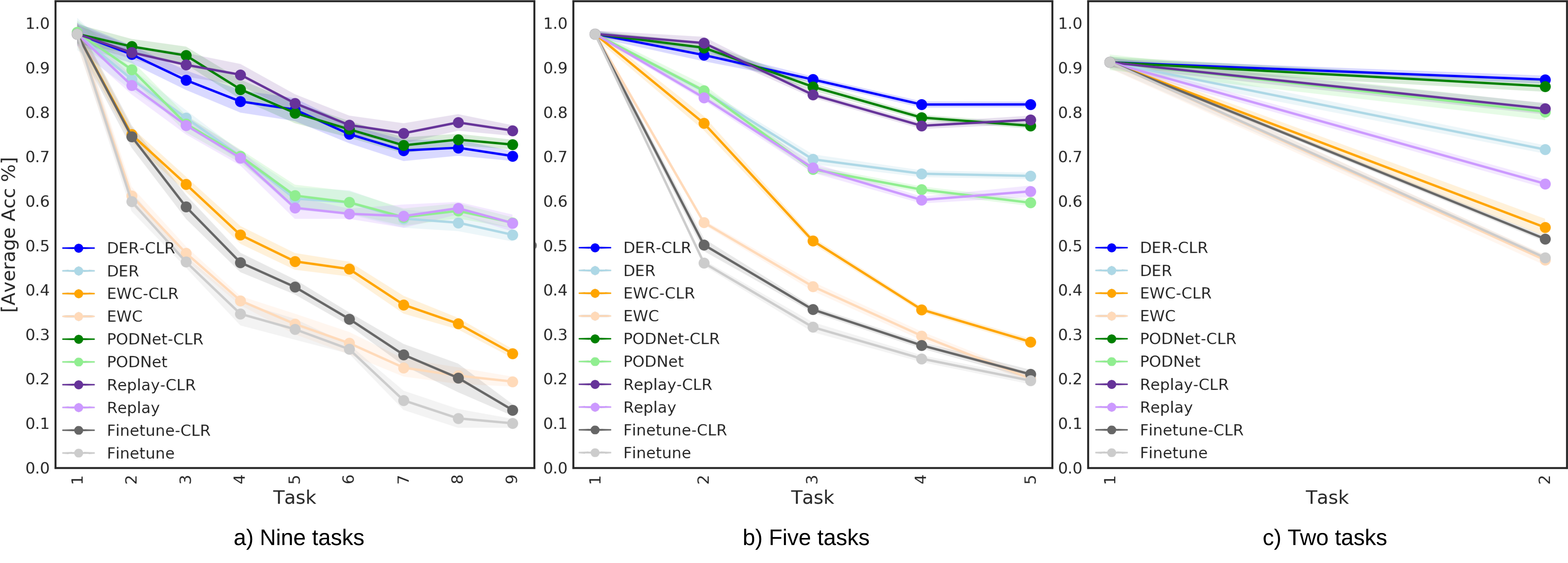}
  \caption{Sub-figures a), b), and c) illustrate the average accuracy for class incremental learning with nine, five, and two tasks, respectively. The baseline model trained on the standard dataset is denoted without the suffix ``CLR", while the model trained on the CL robust dataset is denoted with the suffix ``CLR"  (Best viewed in color).}
  \label{fig:fig_results}
\end{figure*}

\begin{table}[htbp]
\centering
\caption{Presents average accuracy for the CIFAR10 dataset in nine, five, and two task settings. The baseline models suffix with "CLR" trained on a CL robust dataset ($\mathcal{D}_{R}$), while the baseline models without the suffix trained on the standard CIFAR10 dataset ($\mathcal{D}$).}
\begin{tabular}{|c|c|c|c|}
\hline 
\textbf{CL Method}  & \multicolumn{3}{c}{\textbf{Split-CIFAR10}}\vline \\

                & 
                \textbf{Nine Tasks}
                & 
                \textbf{Five Tasks}
                & 
                \textbf{Two Tasks}
                \\ \hline
Multi-Task            &  \multicolumn{3}{c}{94.8$\pm{0.61}$}\vline
\\
\hline 
Finetune                         &
11.11$\pm{0.68}$ &
19.6$\pm{0.81}$ &
45.31$\pm{0.54}$
\\  [0.05cm] 

Finetune-CLR                      &
\textbf{13.0$\pm{0.50}$} &
\textbf{21.03$\pm{1.01}$} &
\textbf{51.48$\pm{0.63}$}
\\  [0.05cm] 

\hline
EWC                         &
19.39 $\pm{1.1}$ &
20.05 $\pm{0.65}$ &
46.73 $\pm{1.16}$ 
\\  [0.05cm] 

EWC-CLR                    &
\textbf{25.71 $\pm{1.0}$ } &
\textbf{28.33 $\pm{0.70}$ } &
\textbf{54.07 $\pm{1.95}$ }
\\  [0.05cm] 

\hline
Replay                         &
55.06 $\pm{1.9}$ &
62.18 $\pm{1.30}$ &
63.9 $\pm{0.97}$ 
\\  [0.05cm] 

Replay-CLR                        &
\textbf{75.86 $\pm{1.2}$ } &
\textbf{78.29 $\pm{0.93}$ } &
\textbf{80.77 $\pm{1.09}$ }
\\  [0.05cm] 
\hline
DER                         &
52.42 $\pm{1.5}$ &
65.65 $\pm{0.72}$ &
71.6 $\pm{0.94}$ 
\\  [0.05cm] 

DER-CLR                    &
\textbf{70.13 $\pm{1.1}$ } &
\textbf{81.74 $\pm{0.65}$ } &
\textbf{87.32 $\pm{0.90}$ }
\\  [0.05cm]

\hline
PODNet                         &
55.14 $\pm{1.2}$ &
59.62 $\pm{0.74}$ &
80.09 $\pm{1.16}$ 
\\  [0.05cm] 

PODNet-CLR                    &
\textbf{72.71 $\pm{1.87}$ } &
\textbf{76.94 $\pm{0.63}$ } &
\textbf{85.82 $\pm{1.23}$ }
\\  [0.05cm]

\hline

\end{tabular}

\label{tab:tab_results}
\end{table}

\section{Experimental Setup}
\label{sec:experimental_setup}
\subsection{Dataset}
We conducted our experiments using the CIFAR10 dataset, a widely used benchmark dataset in CL \cite{deng2022remember, de2021continual, krizhevsky2009learning, kang2022class, de2021continual, yan2021dynamically}. It comprises 60,000 color images categorized into ten classes, each containing 6,000. The dataset is further partitioned into 50,000 training and 10,000 test images, each with a resolution of 32x32 pixels. This dataset is a suitable choice for evaluating the performance of CL models due to its small image size and intra-class variation \cite{de2021continual}.

\subsection{CL Protocol}
We conducted our experiments in the CIL setting \cite{van2019three}, widely regarded as the most challenging scenario in CL. The CIL closely simulates real-world situations where the CL model sequentially learns new classes \cite{de2021continual}. We performed experiments in two, five, and nine class settings to evaluate the effectiveness of $\mathcal{D}_{R}$ in CL.

\subsection{Baseline Models}
We evaluated the six baseline approaches on the standard $\mathcal{D}$ and $\mathcal{D_R}$. The first baseline approach is a replay-based strategy, which stores previous task data and replays it during training. The second approach is the Dynamically Expanding Representation (DER), which increases the network capacity to accommodate new tasks while preserving the previously learned representations \cite{yan2021dynamically}. The third approach is Elastic Weight Consolidation (EWC), a regularization-based approach that adds a penalty term to the loss function that encourages the model to stay close to the critical parameters for previous tasks \cite{kirkpatrick2017overcoming}. The fourth approach is PODNet which utilizes a spatial distillation loss to maintain consistent representations across the learning of the sequential tasks, reducing significant shifts in representation \cite{douillard2020podnet}. The fifth approach is a simple fine-tuning network that does not integrate any particular technique for retaining previous task knowledge. The fine-tuning network is considered the lower bound for the performance of CL. Finally, the multi-task model, which learns all tasks simultaneously and achieves the highest performance, is considered an upper-bound model. To evaluate the impact of the $\mathcal{D_R}$, we trained the aforementioned baseline methods on both the $\mathcal{D}$ and the $\mathcal{D_R}$ datasets in CIL settings with two, five, and nine tasks.

\subsection{Model Training}
In all experiments, we employed the standard residual network architecture \cite{he2016deep}, initialized with random values \cite{glorot2010understanding}. The residual network consists of 34 layers and 21 million parameters. We used the stochastic gradient descent with a learning rate of 0.01 as the optimizer, a momentum of 0.9, and a batch size of 256 used to train all models. The number of training epochs for each task was set to 64. Standard data augmentation techniques were employed during the training phase, including random crops and horizontal flips. We used cosine annealing to gradually decrease the learning rate during each task learning as described in \cite{loshchilov2016sgdr}.

\subsection{Evaluation Matrics}
We use average accuracy as the metric to quantify the performance of the CL models \cite{de2021continual, aljundi2019gradient, liu2020generative, rebuffi2017icarl, kang2022class}; It is the most commonly used metric in CL \cite{de2021continual, lopez2017gradient}. Average accuracy is calculated using the final trained model on all the learned tasks. Average accuracy can be defined as:
\begin{equation}
    ACC = \frac{1}{T} \sum_{i=1}^{T}R_{T,i},
\end{equation}
where $R$ stands for the single model accuracy, $T$ is the total number of tasks, and $i$ represents the task index.

\section{Results and Discussion}
\label{sec:result_and_discussion}
We investigated the role of continually robust features in CL and their impact on mitigating catastrophic forgetting. First, we used the oracle model to construct the $\mathcal{D_R}$. We then compared the performance of corresponding baselines trained on both $\mathcal{D}$ and $\mathcal{D_R}$ in the context of continual incremental learning (CIL) tasks involving two, five, and nine tasks. Overall, our results indicate that the baselines trained on $\mathcal{D_R}$ achieved higher average accuracy than those trained on $\mathcal{D}$, indicating that CL models trained on $\mathcal{D_R}$ are less susceptible to catastrophic forgetting.  We considered six baselines: 1). Replay-based approach, 2). dynamically expanding representation-based approach \cite{yan2021dynamically}, which increases the network capacity to accommodate new tasks while preserving the previously learned representations 3). regularization-based approach, and 4). PODNet, a spatial distillation-based approach 5). fine-tune, and 6). multitask model.

Table \ref{tab:tab_results} and Figure \ref{fig:fig_results} present the final task average accuracy and task-wise average accuracies, respectively. As shown in Table \ref{tab:tab_results}, the replay-based approach trained on the $\mathcal{D_R}$ (Replay-CLR) demonstrated higher average accuracy compared to the replay-based approach trained on the $\mathcal{D}$ across the nine, five, and two tasks in CIL settings. Similarly, the DER-CLR, EWC-CLR, and PODNet-CLR trained on the $\mathcal{D_R}$ demonstrated superior performance to the DER, EWC, and PODNet trained on the $\mathcal{D}$, consistently across the nine, five, and two tasks in CIL settings, respectively. Note that DER-CLR  demonstrated superior utilization of the $\mathcal{D_R}$ compared to Replay-CLR in the five and two task settings, resulting in higher average performance. The PODNet-CLR has obtained comparatively higher average accuracy than DER-CLR in nine task settings. However, Replay-CLR achieved higher average accuracy in the nine-task setting than DER-CLR and PODNet-CLR. The inferior performance of DER-CLR and PODNet-CLR compared to Replay-CLR, particularly in large sequences of tasks, indicates that both DER-CLR and PODNet-CLR could not leverage the $\mathcal{D_R}$ as effectively as the Replay-CLR on the large task sequence. Finally, it is worth noting that the Finetune-CLR model, trained on $\mathcal{D_R}$, also demonstrated superior performance than the Finetune model, trained on $\mathcal{D}$, consistently across all three tasks in the CIL setting.

Our observations highlight that CL's robust features were vital in mitigating catastrophic forgetting. These findings are consistent with similar observations made in the field of adversarial learning \cite{2022adversarially}, where a model trained on robust features has shown greater resilience to adversarial attacks than trained on standard features \cite{ilyas2019adversarial}. It is important to note that we extend the observations made in adversarial robustness to the domain of CL, where a CL model trained on the  $\mathcal{D_R}$ exhibited less forgetting. Furthermore, the findings of this study underline the potential of CL's robust features in improving the robustness of CL models in mitigating catastrophic forgetting. These results suggest that incorporating CL robust features into the training process improves CL models' robustness and mitigates the problem of catastrophic forgetting.

We plan to expand our analysis in several ways. First, we aim to include other models in the experiments to provide a more comprehensive evaluation of the effectiveness of the CL robust features. Second, we aim to perform experiments on datasets with large sequences of tasks, such as CIFAR100, to assess the generalizability of our findings. Third, we aim to analyze the CL robust features learned by different oracle models to understand better their properties and how they contribute to mitigating catastrophic forgetting in CL. Finally, we aim to analyze the $\mathcal{D_R}$ learned by oracle models obtained through various CL strategies, such as  knowledge distillation-based, and dynamic architecture-based approaches.

\section{Conclusion}
\label{sec:conclusion}

We introduced the CL robust dataset and evaluated its effectiveness in preventing catastrophic forgetting in the CIL settings using the CIFAR10 dataset. We demonstrated that the CL model trained on the CL robust dataset achieved a higher average accuracy than when trained on the standard dataset, emphasizing the vital role of robust features in the context of continual learning. This study provides valuable insights into using robust feature extraction for CL and its potential to reduce catastrophic forgetting, thus contributing to a better understanding of CL mechanisms.



\section*{Acknowledgment}
This work was supported by the National Science Foundation Awards NSF ECCS-1903466 and NSF OAC- 2234836. We are also grateful to UK EPSRC support through EP/T013265/1 project NSF-EPSRC: ShiRAS. Towards Safe and Reliable Autonomy in Sensor Driven Systems. Hikmat Khan is partially supported by the US Department of Education through a Graduate Assistance in Areas of National Need (GAANN) program Award Number P200A180055.

\bibliographystyle{unsrt}
\bibliography{main.bib}
\end{document}